\documentclass[conference,letterpaper]{IEEEtran}
\usepackage[utf8]{inputenc}
\usepackage{times,amsmath,amssymb,amsfonts,amsthm}
\usepackage{color}
\usepackage{graphicx}
\graphicspath{ {./Figures/} }

\newtheorem{definition}{Definition}

\theoremstyle{definition}

\newtheorem{remark}{Remark}

\newcommand{\ds}{\displaystyle}
\newcommand{\Real}{\mathbb{R}}

\begin{document}

\newpage
\thispagestyle{empty}
\begin{minipage}{0.8\textwidth}
\noindent {\Huge  IEEE Copyright Notice} \vspace{2cm}

\noindent \copyright 2023 IEEE.  Personal use of this material is permitted.  Permission from IEEE must be obtained for all other uses, in any current or future media, including reprinting/republishing this material for advertising or promotional purposes, creating new collective works, for resale or redistribution to servers or lists, or reuse of any copyrighted component of this work in other works.
\vspace{1cm}

\noindent Accepted to be Published in: Proceedings of the 2023 Latin America Conference on Computational Intelligence (LA-CCI 2023), October 29 - 01 November, 2023, Recife, Brazil.
\end{minipage}

\title{Training Single-Layer Morphological Perceptron Using Convex-Concave Programming\thanks{This work was partially supported by CNPq under grant no. 315820/2021-7 and FAPESP under grant no. 2022/01831-2.}
}

\author{\IEEEauthorblockN{1\textsuperscript{st} Iara Cunha}
\IEEEauthorblockA{Department of Mathematics \\ 
Federal University of Technology – Paraná, \\ 
Ponta Grossa, Brazil. email: iarasilva@utfpr.edu.br}
\and 
\IEEEauthorblockN{2\textsuperscript{nd} Marcos Eduardo Valle}
\IEEEauthorblockA{Department of Applied Mathematics\\ Universidade Estadual de Campinas (Unicamp) \\ Campinas, Brazil. email: valle@ime.unicamp.br}
}

\IEEEoverridecommandlockouts
\IEEEpubid{\makebox[\columnwidth]{979-8-3503-4807-1/23/\$31.00~\copyright2023 IEEE\hfill} 
\hspace{\columnsep}\makebox[\columnwidth]{ }}

\maketitle

\begin{abstract}
This paper concerns the training of a single-layer morphological perceptron using disciplined convex-concave programming (DCCP). We introduce an algorithm referred to as $K$-DDCCP, which combines the existing single-layer morphological perceptron (SLMP) model proposed by Ritter and Urcid with the weighted disciplined convex-concave programming (WDCCP) algorithm by Charisopoulos and Maragos. The proposed training algorithm leverages the disciplined convex-concave procedure (DCCP) and formulates a non-convex optimization problem for binary classification. To tackle this problem, the constraints are expressed as differences of convex functions, enabling the application of the DCCP package. The experimental results confirm the effectiveness of the $K$-DDCCP algorithm in solving binary classification problems. Overall, this work contributes to the field of morphological neural networks by proposing an algorithm that extends the capabilities of the SLMP model. 
\end{abstract}


\section{Introduction}
\label{section0}

Mathematical morphology is a powerful theory of non-linear operators based on geometric and topological concepts \cite{heijmans95,soille99}. From a mathematical point of view, mathematical morphology is well defined using lattice theory \cite{birkhoff93,ronse90}. In a few words, a complete lattice is a partially ordered set in which any subset admits supremum and infimum. The set of the extended real numbers $\bar{\mathbb{R}} = \mathbb{R} \cup \{+\infty,-\infty\}$ is an example of a complete lattice. By enriching $\bar{\mathbb{R}}$ with addition-like operations, we obtain the algebraic structure $(\mathbb{R},\vee,\wedge,+,+')$ from the minimax algebra \cite{cuninghame-green79}. Accordingly, the symbols ``$\vee$'' and ``$\wedge$'' denote the maximum and the minimum operations, respectively. The symbols ``$+$'' and ``$+'$'' coincide with the addition on real numbers and differ only at the infinities. The minimax algebra has been successfully applied, for example, in optimization problems on graphs and machine scheduling \cite{cuninghame-green79,cuninghame-green95}. 

Broadly speaking, the mathematical structure $(\mathbb{R},\vee,\wedge,+,+')$ resembles the real field, but the multiplication is replaced by addition, and the sum is substituted by either the maximum or the minimum operations. Replacing operations of addition and multiplication of real numbers with lattice-based operations of minimax algebras gave birth to the class of morphological networks in the middle 1990s \cite{Davidson1993,Ritter1996,Ritter1998MorphologicalMemories}. The morphological perceptron and the morphological associative memories feature among the first morphological neural networks \cite{ritter96c,Ritter1998MorphologicalMemories}. Despite the algebraic similarity, morphological neural networks are defined as a network whose neurons perform an operation from mathematical morphology \cite{Sussner2011MorphologicalAlgorithm}.

Like the traditional multi-layer perceptron network, a multi-layer morphological perceptron (MLMP) is composed of at least two layers of neurons formulated using lattice-based operations from minimax algebra \cite{ritter96c,ritter97e}. Motivated by the importance of dendritic structures in neuron cells, Ritter and Urcid proposed the so-called dendrite morphological neural networks \cite{Ritter2003,RitterBook}. From a theoretical point of view, a single-layer morphological perceptron (SLMP) with dendrite computation can solve any classification problem in which the classes are compact (see Theorem 1 in \cite{Ritter2003}). 

Besides incorporating dendrite computation on the morphological perceptron, Ritter and Urcid proposed a greedy algorithm for training an SLMP model for binary classification problems. In a few words, the greedy algorithm of Ritter and Urcid adds dendrite terminal fibers until all the training data is correctly classified. In a similar fashion, Sussner and Esmi proposed a greedy algorithm for training morphological perceptrons with competitive learning \cite{Sussner2011MorphologicalAlgorithm}. Despite their computational simplicity, the greedy algorithms are subject to overfitting the training data. In order to avoid overfitted models, morphological neural networks can be trained, for example, using genetic algorithms \cite{Valente2013AAlgorithm,Araujo2012AnEstimation} or solving appropriate optimization problems \cite{charisopoulos17,Oliveira2021LinearProcedure}.

\IEEEpubidadjcol

In contrast to the greedy algorithm proposed by Ritter and Urcid, Charisopoulos and Maragos propose training a single morphological perceptron by solving a convex-concave optimization problem \cite{charisopoulos17}. In this formulation, which resembles the formulation of support vector machines (SVMs) \cite{Cortes1995}, the goal is to minimize the hinge loss classification errors \cite{Rosasco2004AreSame}. However, because of the lattice-based operations, the objective and constraints do not yield a convex optimization problem but a convex-concave programming problem \cite{Yuille2003}. Thus, the training algorithm employed by Charisopoulos and Maragos uses the convex-concave procedure to determine the optimal weight assignment for a binary classification problem \cite{Yuille2003,Shen:2016}. In a similar fashion, we interpret the training of an SLMP model as a convex-concave optimization problem in this paper. Precisely, we train an SLMP model using disciplined convex-concave programming \cite{Shen:2016}. However, unlike the single morphological perceptron model presented in \cite{Maragos:2017}, we consider the general approach of Ritter and Urcid based on dendrite computation. Specifically, we consider a lattice-based network with $K$ fixed dendrites. The proposed training algorithm is referred to as the $K$-dendrite disciplined convex-concave procedure ($K$-DDCCP) because it uses the disciplined convex-concave programming methodology to train a SLMP with $K$ dendrites.



This work is structured as follows:  Section \ref{section1} gives a brief overview of the SLMP model with dendrite computation proposed by Ritter and Urcid \cite{Ritter2003}. This section also presents the WDCCP method proposed by Charisopoulos and Maragos \cite{Maragos:2017}, and it finishes with the proposed non-convex optimization problem. In Section \ref{section2}, we rewrite the optimization problem from the previous section using the difference of convex functions so the training can be addressed using disciplined convex-concave programming \cite{Shen:2016}. The results of the experiments are presented in Section \ref{section3}. The paper finishes with the concluding remarks in Section \ref{section4}.

\section{Single-Layer Morphological Perceptron}
\label{section1}



Advancements in neurobiology and the biophysics of neural computation have highlighted the critical role of dendritic structures in neurons. In a few words, a neuron has dendritic postsynaptic regions that receive input from the terminal branches of other neurons. These terminal branches provide either excitatory or inhibitory input through their terminal buttons, and the postsynaptic membrane of the dendrites determines the excitatory or inhibitory response to the received input. The dendrite structures have been recognized as the primary computational units capable of performing logical operations. In view of this remark, Ritter and Urcid developed a lattice-based neural model with dendritic structures called single-layer morphological perceptron (SLMP) \cite{Ritter2003}. 

The computations of SLMP are performed using the algebraic structure $(\mathbb{R},\vee,\wedge,+,+')$ from minimax algebra \cite{cuninghame-green79}, where $\vee$ and $\wedge$ denote the supremum and infimum operations, respectively. Accordingly, the maximum and the minimum of a finite set of real numbers $X = \{x_1,x_2,\ldots,x_n\} \subseteq \mathbb{R}^n$ are denoted respectively by
\begin{equation}
 \bigvee X \equiv \bigvee_{i=1}^n x_i = \max \{x_i:i=1,\ldots,n\}, 
\end{equation}
and
\begin{equation}
 \bigwedge X \equiv \bigwedge_{i=1}^n x_i = \min \{x_i:i=1,\ldots,n\}.
\end{equation}
The operations ``$+$'' and ``$+'$'' coincide with the usual addition for finite real numbers and differ at the infinities as follows:  
\begin{equation} \label{eq:+}
(+\infty)+(-\infty) = (-\infty)+(+\infty) = -\infty,
\end{equation} 
and 
\begin{equation} \label{eq:+'}
(+\infty)+'(-\infty) = (-\infty)+'(+\infty) = +\infty. 
\end{equation}
In the following, we only consider finite values. Thus, we do not need to distinguish between ``$+$'' and ``$+'$'', and we shall consider only the traditional addition ``$+$'' of real numbers. Detailed accounts on the mathematical structure $(\mathbb{R},\vee,\wedge,+,+')$ can be found, for example, in \cite{cuninghame-green79,Sussner2011MorphologicalAlgorithm}.

In mathematical terms, suppose a single neuron with $K$ dendrites receives an input $\boldsymbol{x} = [x_1,x_2,\ldots,x_n] \in \mathbb{R}^n$. Using lattice-based operations, the computation performed by the $k$th dendrite, for $k=1,\ldots,K$, is given by 
\begin{equation}
      \tau_k(\boldsymbol{x}) = \displaystyle  p_k \bigwedge_{i=1}^n \bigwedge_{l \in \{0,1\}} (-1)^{1-l} (x_i+w_{ik}^l), 
     \label{eq:MNNmin}
 \end{equation}
where $w_{ik}^l$ denotes the weight of the dendrite fiber coming from the $i$th input neuron into $k$th dendrite and $p_k \in \{-1,+1\}$ indicates its role, with $+1$ and $-1$ for excitatory and inhibitory junctions, respectively.
The state of a neuron is determined by a function considering the inputs received from all of its dendrites. Formally, the output of a neuron with dendrite computation is given by 
\begin{equation} \label{eq:tau_min}
\tau(\boldsymbol{x}) = \bigwedge_{k=1}^K \tau_k(\boldsymbol{x}),
\end{equation}
and its state is determined by means of the equation
\begin{equation}
\label{y_slmp}
y(\boldsymbol{x}) = f \left ( \bigwedge_{k=1}^K  \left [ \displaystyle  p_k \bigwedge_{i=1}^n \bigwedge_{l \in \{0,1\}} (-1)^{1-l}(x_i+w_{ik}^l) \right ]  \right ),
\end{equation}
where the hard limiter activation function given below is used:
\begin{equation}
\label{hard_limiter}
f(\tau) = 
\begin{cases}
   0, & \text{if } \tau \geq 0, \\
   1, & \text{if } \tau < 0.
\end{cases}
\end{equation}

\begin{remark}
Lattice theory is endowed with a duality principle \cite{birkhoff93,cuninghame-green79}. In a few words, the duality principle asserts that every statement corresponds to a dual one obtained by interchanging the operations ``$\vee$ and ``$\wedge$'', and vice-versa. Accordingly, by replacing the minimum with the maximum operation, the computation performed by the $k$th dendrite can be expressed as follows
\begin{equation} \label{eq:MNNmax}
     \tau_k(\boldsymbol{x}) = \displaystyle  p_k \bigvee_{i=1}^n \bigvee_{l \in \{0,1\}} (-1)^{1-l} (x_i+w_{ik}^l).
 \end{equation}
Similarly, the output of a neuron with $K$ dendrites can be computed by means of the equation
\begin{equation} \label{eq:tau_max}
\tau(\boldsymbol{x}) = \bigvee_{k=1}^K \tau_k(\boldsymbol{x}),
\end{equation}
and its state is given by $y(\boldsymbol{x}) = f(\tau(\boldsymbol{x}))$.
We would like to point out that one can move from \eqref{eq:MNNmax} and \eqref{eq:MNNmin} as well as from \eqref{eq:tau_max} and \eqref{eq:tau_min} by redefining the weights and/or changing $p_k$ from excitatory to inhibitory, and vice-versa.
\end{remark}

Let us conclude by pointing out that Ritter and Urcid proposed an algorithm for training an SLMP for binary classification problems \cite{Ritter2003}. Their algorithm behaves as a greedy heuristic, where the dendrites of the morphological neuron grow gradually as long as there are misclassified training patterns. The training process only ends when all the training data is correctly classified. Consequently, it is possible that the morphological neural network overfits the dataset.

\subsection{Weighted Disciplined Convex-Concave Programming}

In contrast to the greedy training algorithm of Ritter and Urcid, Charisopoulos and Maragos trained a single morphological perceptron by solving an optimization problem \cite{Maragos:2017}, in a similar fashion to traditional support vector machines \cite{Cortes1995}. By formulating the training as an optimization problem, one obtains a convex cost function, but the constraints are composed of inequalities given by the difference of convex functions (DC) or, equivalently, convex-concave functions. The following reviews the training method proposed by Charisopoulos and Maragos. 

First of all, Charisopoulous and Maragos considered a network with a single morphological neuron described by the equation
\begin{equation} \label{eq:single-perceptron}
y(\boldsymbol{x}) = f\left(\bigvee_{i=1}^n x_i + w_i \right),
\end{equation}
for all input $\boldsymbol{x} = [x_1,\ldots,x_n] \in \mathbb{R}^n$. We would like to remark that \eqref{eq:single-perceptron} corresponds to a particular SLMP with only one excitatory dendrite ($K=1$ and $p_1 = +1$) and weights satisfying $w_{i1}^0 = -\infty$ and $w_{i1}^1 \equiv w_i$, for all $i=1,\ldots,n$.

Consider a training set $\mathcal{T} = \{(\boldsymbol{x}^{(j)},y^{(j)}):j=1,\ldots,M\} \subseteq \mathbb{R}^n \times \{c_0,c_1\}$, where $c_0$ and $c_1$ represent the class labels in a binary classification problem. Define the sets 
\begin{equation} \label{eq:sets}
C_0 = \{\boldsymbol{x}^{(j)}:y^{(j)} = c_0 \} \quad \text{and} \quad C_1 = \{\boldsymbol{x}^{(j)}:y^{(j)} = c_1 \},
\end{equation}
of the training samples associated with the labels $c_0$ and $c_1$, respectively.
Charisopoulos and Maragos proposed the 
following optimization problem for training the morphological model given by \eqref{eq:single-perceptron}:
\begin{equation}
\begin{array}{rrclcl} 
\ds \min_{\boldsymbol{w},\boldsymbol{\xi}} &  \multicolumn{3}{l}{\ds \sum_{j = 1}^M \nu_j \max (\xi_j, 0),} \\
\textrm{s.t.} & \ds  \bigvee_{i=1}^n  
  (x_i^{(j)}+w_{i}) & \leq & \xi_j, & \mbox{ if }  \boldsymbol{x}^{(j)} \in C_0,\\
&\ds   \bigvee_{j=1}^n  
 (x_{i}^{(j)}+w_{i}) & \geq & - \xi_j,  & \mbox{ if }  \boldsymbol{x}^{(j)}  \in C_1, \\
\end{array}
\label{eq:MaragosModel}
\end{equation}
where $\boldsymbol{w}=[w_1,\ldots,w_n] \in \Real^n$ is a weight vector and $\boldsymbol{\xi}=[\xi_1,\ldots,\xi_M] \in \Real^M$ is the slack variable vector used to ensure that only misclassified patterns contribute to the objective (loss) function. Furthermore, the weights $\nu_j$ are introduced into the objective function to penalize patterns with a greater likelihood of being outliers. Because of the maximum operations, the constraints in the optimization problem \eqref{eq:MaragosModel} are expressed by the difference between convex and concave functions, resulting in a convex-concave optimization problem \cite{Yuille2003}. The training algorithm based on the convex-concave procedure proposed by Charisopoulos and Maragos can be implemented using the DCCP library for the CVXPY, an open-source optimization package for \texttt{python} \cite{Shen:2016,Diamond2016cvxpy}.

Besides the morphological network given by \eqref{eq:single-perceptron}, Charisopoulos and Maragos also considered a model given by the convex combination of two morphological neurons as follows $\lambda \in [0, 1]$.
\begin{equation}
y(\boldsymbol{x}) = f \left( \lambda \left ( \bigvee_{i=1}^n x_i+w_i \right ) + (1-\lambda) \left ( \bigwedge_{i=1}^n x_i+m_i \right ) \right).
\label{eq:MaragosResponse}
\end{equation}


\section{Training a Single-Layer Morphological Perceptron Using Convex-Concave Programming}
\label{section2}

We propose a training algorithm for a single-layer morphological perceptron featuring K dendrites. Drawing inspiration from the Charisopoulos and Maragos approach, our training method for an SLMP network aims to minimize the slack variables within the constraints established by the decision functions presented in Ritter and Urcid's SLMP model, as detailed in \cite{Ritter2003}. Precisely, given a training set $\mathcal{T} = \{(\boldsymbol{x}^{(j)},y^{(j)}):j=1,\ldots,M\} \subseteq \mathbb{R}^n \times \{c_0,c_1\}$, the weights of an SLMP are obtained by solving the optimization problem:
\begin{equation}
\begin{array}{rrclcl} 
\ds \min_{\boldsymbol{W}, \boldsymbol{\xi} } &  \multicolumn{3}{l}{\ds \sum_{j = 1}^M \max (\xi_j, 0)} \\
\textrm{s.t.} & \ds \tau(\boldsymbol{x}^{(j)}) & \leq & \xi_j, & \mbox{ if }  \boldsymbol{x}^{(j)} \in C_0,\\
&\ds \tau(\boldsymbol{x}^{(j)}) & \geq & - \xi_j,  & \mbox{ if }  \boldsymbol{x}^{(j)}  \in C_1, \\
\end{array}
\label{eq:NayveModel}
\end{equation}
where $\tau(\boldsymbol{x}^{(j})$ denotes the output of a neuron with dendrite computation under presentation of $\boldsymbol{x}^{(j)}$, and $C_0$ and $C_1$ are given by \eqref{eq:sets}. Unfortunately, unlike \eqref{eq:MaragosModel}, \eqref{eq:NayveModel} is not a convex-concave optimization problem and we need additional assumptions and some mathematics to solve it.

\subsection{Geometric Interpretation and Additional Assumptions}

In this subsection, we provide a geometric interpretation of the SLMP that supports additional assumptions considered to solve \eqref{eq:NayveModel}.

First of all, the output of the $k$th dendrite given by \eqref{eq:MNNmin} can be expressed as follows for all $\boldsymbol{x} = [x_1,\ldots,x_n]\in \mathbb{R}^n$:
\begin{equation}
    \tau_k(\boldsymbol{x}) = \bigwedge_{i=1}^n(x_i+w^1_{ik})\wedge  \bigwedge_{i=1}^n(-x_i-w^0_{ik}).
\end{equation}
Note that $\tau_k(\boldsymbol{x})\geq 0$ if and only if $-w^1_{ik} \leq x_i \leq -w^0_{ik}$, for all $i=1,\ldots,n$. Therefore, the inequality $\tau_k(\boldsymbol{x}) \geq 0$ holds if and only if the pattern $\boldsymbol{x}$ belongs to the hyperbox whose bottom-left and top-right corners are $- \boldsymbol{w}^1_k =  [-w^1_{1k},\ldots,-w^1_{nk}] \in \mathbb{R}^n$ and $ - \boldsymbol{w}^0_k = [-w^0_{1k},\ldots,-w^0_{nk}]  \in \mathbb{R}^n$, respectively. In other words, a dendrite determines a hyperbox in $\Real^n$. By considering only excitatory dendrites, the output of a neuron given by \eqref{eq:tau_max} satisfies $\tau(\boldsymbol{x}) \geq 0$ if and only if $\boldsymbol{x}$ belongs to the hyperbox determined by at least one dendrite. In other words, the decision surface of a morphological neuron with dendrite computation corresponds to the union of the hyperbox determined by its dendrites if $p_k=+1$ for all $k=1,\ldots,K$. 


Consider an SLMP with $K$ excitatory dendrites, that is, $p_k=+1$ for all $k = 1,\ldots,K$. Given a training set $\mathcal{T} = \{(\boldsymbol{x}^{(j)},y^{(j)}):j=1,\ldots,M\}$, the weights $\boldsymbol{w}_k = [\boldsymbol{w}_k^1,-\boldsymbol{w}_k^0] \in \Real^N$, where $N=2n$, obtained by concatenating $\boldsymbol{w}_k^1$ and $-\boldsymbol{w}_k^0$, can be determined 
by solving the following optimization problem:

\begin{equation}
\begin{array}{rrclcl} 
\ds \min_{\boldsymbol{W}, \boldsymbol{\xi}} &  \multicolumn{3}{l}{\ds \sum_{j = 1}^M \max (\xi_j, 0)} \\
\textrm{s.t.} & \ds \bigvee_{k=1}^K \bigwedge_{i=1}^N  
  \left [ z_i^{(j)}+(w_k)_i \right ]& \leq & \xi_j & \mbox{ if }  \boldsymbol{z}^{(j)} \in C_0\\
&\ds \bigvee_{k=1}^K \bigwedge_{j=1}^N  
 \left [ z_i^{(j)}+(w_k)_i \right ] & \geq & - \xi_j  & \mbox{ if }  \boldsymbol{z}^{(j)}  \in C_1 \\
\end{array}
\label{eq:mainModel}
\end{equation}
where $\boldsymbol{z}^{(j)}=[\boldsymbol{x}^{(j)},-\boldsymbol{x}^{(j)}] \in \Real^N$ is obtained by concatenating $\boldsymbol{x}^{(j)}$ and $-\boldsymbol{x}^{(j)}$ and $\boldsymbol{W} \in \Real^{K \times N}$ is a matrix with its rows being represented by $\boldsymbol{w}_k$.
When considering one dendrite ($K=1$), we encounter the optimization problem presented by Charisopoulos and Maragos in \cite{Maragos:2017}. This problem is a \textit{disciplined convex-concave program} (DCCP) and can be solved using the DCCP library for the \texttt{python}'s CVXPY package \cite{Shen:2016}. However, the problem fails to be DCCP when dealing with more than one dendrite. Thus, the constraints should be expressed as differences of convex functions, which will be further detailed in the following subsection.

\subsection{Alternative Formulation a SLMP Network}

The optimization model defined by \eqref{eq:mainModel} is not a DCCP problem \cite{Shen:2016}. Hence, we must rewrite the constraints in \eqref{mainModel_DC} as the difference of two convex functions (DC). For this purpose, we use the Definition \ref{def:DC} below and used an alternative formulation based on \cite{Bagirov:2020}.

\begin{definition}
\label{def:DC}
A function real-valued function $f$ is called DC on a convex set $S \subseteq \mathbb{R}$ if there exist two convex function $f_1$, $f_2: S  \rightarrow \Real$ such that $f(\boldsymbol{x}) = f_1(\boldsymbol{x})-f_2(\boldsymbol{x})$ for all $\boldsymbol{x} \in S$.
\end{definition}

Consider the following function
\begin{equation}
\varphi(\boldsymbol{z},\boldsymbol{W}) = \bigvee_{k=1}^K \bigwedge_{i=1}^N \left [ z_{i} + (w_{k})_i \right].
\end{equation}
Alternatively, we can write $\varphi$ as follows:
\begin{align}
\varphi(\boldsymbol{z},\boldsymbol{W}) &= \bigvee_{k=1}^K \bigwedge_{i=1}^N \left [ z_{i} + (w_{k})_i \right] \\
&= \bigvee_{k=1}^K \left [  - \left(\bigvee_{i=1}^N \left [ -z_{i} - (w_{k})_i \right]\right) \right]\\
&= \bigvee_{k=1}^K \left [ - \Psi^{k}(\boldsymbol{z},\boldsymbol{w}_k) \right],
\end{align}
where $\Psi^{k}(\boldsymbol{z},\boldsymbol{w}_k) = \ds \max_{i=1,\ldots,N} \{-z_{i} - (w_{k})_i \}$. 
Thus, we have
\begin{equation}
\varphi (\boldsymbol{z},\boldsymbol{W}) = \max_{k = 1, \ldots, K} \{- \Psi^{k}(\boldsymbol{z},\boldsymbol{w}_k) \}.
\end{equation}
Moreover, the function above can be rewritten as
\begin{equation}
\varphi (\boldsymbol{z},\boldsymbol{W}) = \varphi_{1} (\boldsymbol{z},\boldsymbol{W}) - \varphi_{2} (\boldsymbol{z},\boldsymbol{W}),
\end{equation}
where
\begin{equation}
\varphi_{1} (\boldsymbol{z},\boldsymbol{W}) = \ds \max_{k = 1, \ldots, K} \left\{ \sum_{\substack{t=1 \\ t \neq k}}^{K} \Psi^{t}(\boldsymbol{z},\boldsymbol{w}_t) \right\}, 
\end{equation}
and
\begin{equation}
\varphi_{2} (\boldsymbol{z},\boldsymbol{W}) = \ds \sum_{k=1}^{K} \Psi^{k}(\boldsymbol{z},\boldsymbol{w}_k),
\end{equation}
are both functions convex functions.

Concluding, problem \eqref{eq:mainModel} can we rewritten using DC functions as follows:
\begin{equation}
\label{mainModel_DC}
\begin{array}{rrlll} 
\ds \min_{\boldsymbol{W}, \boldsymbol{\xi}} &  \multicolumn{3}{l}{\ds \sum_{j = 1}^M \max (\xi_j, 0)} \\
\textrm{s.t.} & \varphi_{1}(\boldsymbol{z}^{(j)},\boldsymbol{W}) - \varphi_{2}(\boldsymbol{z}^{(j)},\boldsymbol{W}) & \leq \xi_j, & \boldsymbol{z}^{(j)} \in C_0,\\
& \varphi_{1}(\boldsymbol{z}^{(j)},\boldsymbol{W}) - \varphi_{2}(\boldsymbol{z}^{(j)},\boldsymbol{W})  & \geq - \xi_j,  & \boldsymbol{z}^{(j)}  \in C_1. \\
\end{array}
\end{equation}

The problem \eqref{mainModel_DC}, which is equivalent to problem \eqref{eq:mainModel}, can be solved in a straightforward manner using the disciplined convex-concave programming (DCCP) extension package for \texttt{CVXPY} library for \texttt{python} \cite{Shen:2016}.

\section{Computational Experiments}
\label{section3}

In our experiments, we used four databases: Ripley's dataset \cite{Ripley1995}, the synthetic blobs and double moons datasets, and the breast cancer dataset for classification (the three latter are available at \texttt{scikit-learn} \cite{scikit_learn}). The first three datasets have two features (that is, $\boldsymbol{x}^{(j)} \in \Real^2$), whereas the breast cancer dataset has 30 features (e.g. $\boldsymbol{x}^{(j)} \in \Real^{30}$).

\begin{figure*}[t]
\begin{tabular}{cc}
a) SLMP with greedy training. Accuracy: 0.78. &
b) Morphological Perceptron with WDCCP. Accuracy: 0.876 \\
\includegraphics[width=0.42\textwidth]{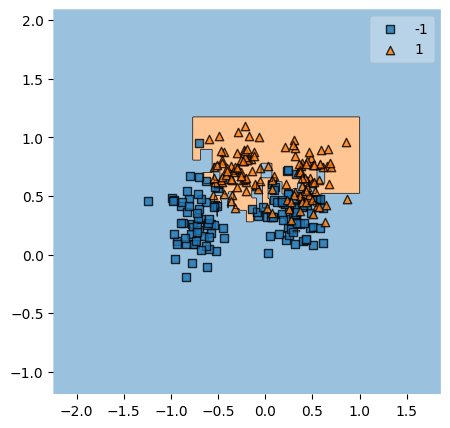} & \includegraphics[width=0.42\textwidth]{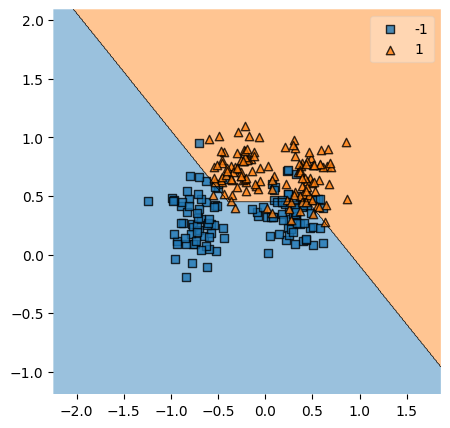} \\
c) Linear SVM. Accuracy: 0.856 & 
d) SLMP with 4-DDCCP. Accuracy: 0.872 \\
\includegraphics[width=0.42\textwidth]{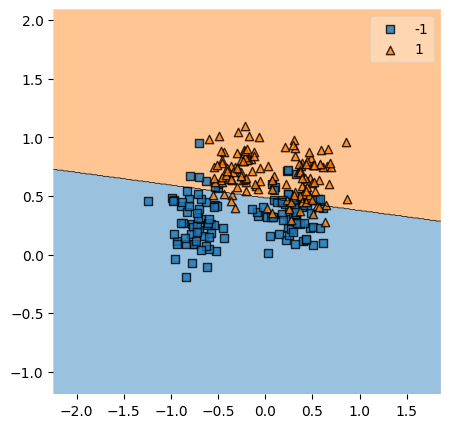} &
\includegraphics[width=0.42\textwidth]{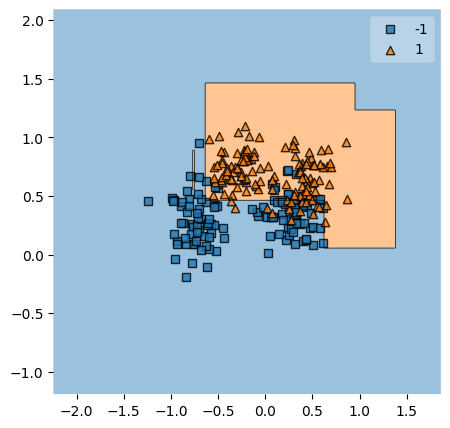}
\end{tabular}
\caption{Decision surface of classifiers and scatter plot of the test set of the Ripley dataset.}
\label{fig:decision_surface}
\end{figure*}

For comparison purposes, we have considered the greedy training algorithm proposed by Ritter and Urcid \cite{Ritter2003}, the WDCCP method proposed by Charisoupoulos and Maragos \cite{Maragos:2017}, and the linear support vector machines (Linear SVM) \cite{Cortes1995Support-vectorNetworks}. We used 1000 training samples and 250 testing samples for the datasets containing two features. For the breast cancer dataset, we used 381 samples for training, while the remaining 188 were used for testing. Moreover, we repeated each experiment 30 times to better compare the accuracy score between the classifiers. Table \ref{tab:score_std_test} summarizes the results of the experiments conducted in this study.
Figure \ref{fig:decision_surface} shows the decision surfaces and the corresponding accuracy scores produced by the morphological networks trained with the greedy algorithm of Ritter and Urcid, the WDCCP, and the proposed 4-DDCCP on the Ripley dataset. This figure also includes the decision surface produced by the linear SVM classifier. 

\begin{table*}[t]
\begin{center}
\caption{Accuracy's mean and standard deviations and the best score.}
\label{tab:score_std_test}
\begin{tabular}{c|cccccc}
& \textbf{Greedy} & \textbf{WDCCP}  & \textbf{2-DDCCP} & \textbf{3-DDCCP} & \textbf{4-DDCCP} & \textbf{SVM} \\ \hline
\textbf{Ripley} & 0.793$\pm$0.007 & \textbf{0.871}$\pm$0.006  & 0.852$\pm$0.015 & 0.854$\pm$0.015 & 0.840$\pm$0.060 & 0.856$\pm$0.000\\
Best: & 0.804 & \textbf{0.876} & \textbf{0.876} & 0.872 & 0.872 & 0.856 \\
\hline 
\textbf{Blobs} & \textbf{0.968}$\pm$0.004 & 0.700$\pm$0.000 & 0.848$\pm$0.195 & 0.932$\pm$0.131 & 0.927$\pm$0.142 & 0.752$\pm$0.000\\
Best: & 0.972 & 0.700 & \textbf{0.988} & \textbf{0.988} & \textbf{0.988} & 0.752 \\
\hline 
\textbf{Double Moons} & \textbf{0.986}$\pm$0.004 & 0.678$\pm$0.002 & 0.885$\pm$0.055 & 0.894$\pm$0.057 & 0.908$\pm$0.033 & 0.880$\pm$0.000\\
Best: & 0.988 & 0.680 & 0.936 & 0.936 & \textbf{0.992} & 0.880 \\
\hline 
\textbf{Breast Cancer} & 0.828$\pm$0.002 & 0.599$\pm$0.008 & 0.940$\pm$0.007 & 0.946$\pm$0.002 & 0.947$\pm$0.000 & \textbf{0.957}$\pm$0.000\\
Best: & 0.830 & 0.606 & 0.947 & 0.947 & 0.947 & \textbf{0.957}\\
\end{tabular}
\end{center}
\end{table*}

Note that the linear SVM and morphological models trained using the WDCCP and $K$-DDCCP algorithms exhibited comparable performance regarding accuracy and robustness for the Ripley dataset. As expected, the greedy training algorithm overfitted the dataset, yielding a 100\% accuracy score for the training set. In contrast, in the blobs dataset, the SLMP with the $K$-DDCCP method outperformed both the SVM method and the morphological network trained using the WDCCP. On the downside, we observed that the $K$-DDCCP method lacked robustness for the blobs dataset due to its substantial standard deviation. The lack of robustness is partially caused by the randomness of the initialization of the method used for solving the convex-concave optimization problem. For the breast cancer dataset, the SLMP with $K$-DDCCP method achieved excellent results, demonstrating robustness due to its low standard deviation.

\section{Concluding Remarks}
\label{section4}

In this paper, we presented the $K$-DDCCP algorithm for training a single-layer morphological perceptron (SLMP). The algorithm extends the capabilities of the SLMP model proposed by Ritter and Urcid by using a training algorithm that uses a convex-concave procedure to solve an optimization problem that resembles the one solved by traditional SVM for binary classification tasks.
Precisely, we formulated a non-convex optimization problem for binary classification using the decision functions defined by Ritter and Urcid. By expressing the constraints as differences of convex functions, we were able to solve it using the DCCP package for \texttt{python}'s \texttt{CVXPY library} \cite{Shen:2016}. 

The experimental results presented in Section \ref{section3} showed that SLMP with the $K$-DDCCP algorithm outperformed the model with the greedy algorithm of Ritter and Urcid in terms of the classification accuracy score. Indeed, the algorithm achieved better generalization performance by incorporating multiple dendrites and utilizing the DCCP framework. This highlights the importance of dendritic structures in neural networks and their ability to perform logical operations.

Concluding, the proposed algorithm contributes to the field of morphological neural networks by providing a novel approach to training single-layer perceptrons with dendritic structures. The use of disciplined convex-concave programming enables the optimization of non-convex problems, extending the applicability of morphological neural networks to complex classification tasks.

Future work could focus on further improving the $K$-DDCCP algorithm by incorporating additional optimization techniques or exploring different variations of morphological neural networks. Additionally, investigating the performance of the algorithm on larger datasets and comparing it with other state-of-the-art classification algorithms would provide a more comprehensive evaluation of its capabilities.

Overall, this work contributes to advancing morphological neural networks and demonstrates the potential of incorporating dendritic structures in designing efficient and effective neural models for classification tasks.


\begin{thebibliography}{10}

\bibitem{heijmans95}
H.~J. A.~M. Heijmans, ``{Mathematical Morphology: A Modern Approach in Image
  Processing Based on Algebra and Geometry},'' {\em SIAM Review}, vol.~37,
  no.~1, pp.~1--36, 1995.

\bibitem{soille99}
P.~Soille, {\em {Morphological Image Analysis}}.
\newblock Berlin: Springer Verlag, 1999.

\bibitem{birkhoff93}
G.~Birkhoff, {\em {Lattice Theory}}.
\newblock Providence: American Mathematical Society, 3~ed., 1993.

\bibitem{ronse90}
C.~Ronse, ``{Why Mathematical Morphology Needs Complete Lattices},'' {\em
  Signal Processing}, vol.~21, no.~2, pp.~129--154, 1990.

\bibitem{cuninghame-green79}
R.~Cuninghame-Green, {\em {Minimax Algebra: Lecture Notes in Economics and
  Mathematical Systems 166}}.
\newblock New York: Springer-Verlag, 1979.

\bibitem{cuninghame-green95}
R.~Cuninghame-Green, ``{Minimax Algebra and Applications},'' in {\em Advances
  in Imaging and Electron Physics} (P.~Hawkes, ed.), vol.~90, pp.~1--121, New
  York, NY: Academic Press, 1995.

\bibitem{Davidson1993}
J.~L. Davidson and F.~A. Hummer, ``Morphology neural networks: An introduction
  with applications,'' {\em Circuits, Systems and Signal Processing}, vol.~12,
  pp.~177--210, 1993.

\bibitem{Ritter1996}
G.~X. Ritter and P.~Sussner, ``An introduction to morphological neural
  networks,'' {\em Proceedings of 13th International Conference on Pattern
  Recognition}, vol.~4, pp.~709--717 vol.4, 1996.

\bibitem{Ritter1998MorphologicalMemories}
G.~X. Ritter, P.~Sussner, and J.~L. Diaz-De-Leon, ``{Morphological associative
  memories},'' {\em IEEE Transactions on Neural Networks}, vol.~9, no.~2,
  pp.~281--293, 1998.

\bibitem{ritter96c}
G.~X. Ritter and P.~Sussner, ``{An Introduction to Morphological Neural
  Networks},'' in {\em Proceedings of the 13th International Conference on
  Pattern Recognition}, (Vienna, Austria), pp.~709--717, 1996.

\bibitem{Sussner2011MorphologicalAlgorithm}
P.~Sussner and E.~L. Esmi, ``{Morphological perceptrons with competitive
  learning: Lattice-theoretical framework and constructive learning
  algorithm},'' {\em Information Sciences}, vol.~181, pp.~1929--1950, 5 2011.

\bibitem{ritter97e}
G.~X. Ritter and P.~Sussner, ``{Associative Memories Based on Lattice
  Algebra},'' in {\em Computational Cybernetics and Simulation}, (Orlando,
  Florida), 1997 IEEE International Conference on Systems, Man, and
  Cybernetics, 1997.

\bibitem{Ritter2003}
G.~Ritter and G.~Urcid, ``{Lattice Algebra Approach to Single-Neuron
  Computation},'' {\em IEEE Transactions on Neural Networks}, vol.~14, no.~2,
  pp.~282--295, 2003.

\bibitem{RitterBook}
G.~X. Ritter and G.Urcid, {\em Introduction to Lattice Algebra: With
  Applications in AI, Pattern Recognition, Image Analysis, and Biomimetic
  Neural Networks}.
\newblock CRC Press, 2021.

\bibitem{Valente2013AAlgorithm}
R.~A. Valente and M.~E. Valle, ``{A Brief Account on Morphological Perceptron
  with Competitive Layer Trained by a Certain Genetic Algorithm},'' in {\em 1st
  BRICS Countries Congress (BRICS-CCI) and 11th Brazilian Congress (CBIC) on
  Computational Intelligence}, (Porto de Galinhas), Sociedade Brasileira de
  Intelig{\^{e}}ncia Computacional, 9 2013.

\bibitem{Araujo2012AnEstimation}
R.~d.~A. Ara{\'{u}}jo, A.~L. Oliveira, S.~Soares, and S.~Meira, ``{An
  evolutionary morphological approach for software development cost
  estimation},'' {\em Neural Networks}, vol.~32, pp.~285--291, 8 2012.

\bibitem{charisopoulos17}
V.~Charisopoulos and P.~Maragos, ``{Morphological Perceptrons: Geometry and
  Training Algorithms},'' in {\em Mathematical Morphology and Its Applications
  to Signal and Image Processing} (J.~Angulo, S.~Velasco-Forero, and F.~Meyer,
  eds.), (Cham), pp.~3--15, Springer International Publishing, 2017.

\bibitem{Oliveira2021LinearProcedure}
A.~L. Oliveira and M.~E. Valle, ``{Linear Dilation-Erosion Perceptron Trained
  Using a Convex-Concave Procedure},'' in {\em In: Abraham A. et al. (eds)
  Proceedings of the 12th International Conference on Soft Computing and
  Pattern Recognition (SoCPaR 2020). SoCPaR 2020. Advances in Intelligent
  Systems and Computing}, vol.~1383, pp.~245--255, Springer, Cham, 12 2021.

\bibitem{Cortes1995}
C.~Cortes and V.~Vapnik, ``Support-vector networks,'' {\em Machine learning},
  vol.~20, no.~3, pp.~273--297, 1995.

\bibitem{Rosasco2004AreSame}
L.~Rosasco, E.~D. Vito, A.~Caponnetto, M.~Piana, and A.~Verri, ``{Are Loss
  Functions All the Same?},'' {\em Neural Computation}, vol.~16,
  pp.~1063--1076, 5 2004.

\bibitem{Yuille2003}
A.~L. Yuille and A.~Rangarajan, ``The concave-convex procedure,'' {\em Neural
  Computation}, vol.~15, no.~4, pp.~915--936, 2003.

\bibitem{Shen:2016}
X.~Shen, S.~Diamond, Y.~Gu, and S.~Boyd, ``Disciplined convex-concave
  programming,'' 2016.

\bibitem{Maragos:2017}
V.~Charisopoulos and P.~Maragos, ``Morphological perceptrons: Geometry and
  training algorithms,'' in {\em Mathematical Morphology and Its Applications
  to Signal and Image Processing}, pp.~3--15, Springer International
  Publishing, 2017.

\bibitem{Diamond2016cvxpy}
S.~Diamond and S.~Boyd, ``Cvxpy: A python-embedded modeling language for convex
  optimization,'' 2016.

\bibitem{Bagirov:2020}
A.~Bagirov, S.~Taheri, N.~Karmitsa, N.~Sultanova, and S.~Asadi, ``Robust
  piecewise linear l 1 -regression via nonsmooth dc optimization,'' {\em
  Optimization Methods and Software}, vol.~37, pp.~1--21, 12 2020.

\bibitem{Ripley1995}
B.~D. Ripley and N.~L. Hjort, {\em Pattern Recognition and Neural Networks}.
\newblock USA: Cambridge University Press, 1st~ed., 1995.

\bibitem{scikit_learn}
F.~{Pedregosa}, G.~{Varoquaux}, A.~{Gramfort}, V.~{Michel}, B.~{Thirion},
  O.~{Grisel}, M.~{Blondel}, P.~{Prettenhofer}, R.~{Weiss}, V.~{Dubourg},
  J.~{Vanderplas}, A.~{Passos}, D.~{Cournapeau}, M.~{Brucher}, M.~{Perrot}, and
  E.~{Duchesnay}, ``Scikit-learn: Machine learning in python.'' Journal of
  Machine Learning Research, 12(Oct):2825-2830, 2011.

\bibitem{Cortes1995Support-vectorNetworks}
C.~Cortes and V.~Vapnik, ``{Support-vector networks},'' {\em Machine Learning},
  vol.~20, pp.~273--297, 9 1995.

\end{thebibliography}

\end{document}